\documentclass[11pt,a4paper]{article}
\usepackage[hyperref]{emnlp2020}
\usepackage{times}
\usepackage{latexsym}

\usepackage{microtype}

\aclfinalcopy % Uncomment this line for the final submission
\def\aclpaperid{1694} %  Enter the acl Paper ID here

\usepackage{url}

\usepackage{amsmath,bm} %add
\usepackage{graphicx}  %add
\def\argmax{\mathop{\rm argmax}}%
\def\argmin{\mathop{\rm argmin}}%
\usepackage{multirow}
\usepackage{diagbox}
\usepackage{amsfonts}
\usepackage{enumitem}
\usepackage{graphics}
\usepackage{adjustbox}
\usepackage{subcaption}

\usepackage{url}
\usepackage{algpseudocode}

\title{Accurate Word Alignment Induction from Neural Machine Translation}

\author{
 Yun Chen\thanks{\:\:Corresponding author. Part of the work was done when Yun was in Huawei Noah’s Ark Lab.} $^{\:1}$,
 Yang Liu$^{2}$, Guanhua Chen$^3$, Xin Jiang$^4$, Qun Liu$^4$\\
 $^1$Shanghai University of Finance and Economics, Shanghai, China\\
 $^2$Tsinghua University, Beijing, China \\
 $^3$The University of Hong Kong, Hong Kong, China \\
 $^4$Huawei Noah's Ark Lab, Hong Kong, China \\
 yunchen@sufe.edu.cn, liuyang2011@tsinghua.edu.cn, \\ 
 ghchen@eee.hku.hk, \{jiang.xin, qun.liu\}@huawei.com\\
}
\date{}

\begin{document}
\maketitle
\begin{abstract}
Despite its original goal to jointly learn to align and translate, prior researches suggest that Transformer captures poor word alignments through its attention mechanism. In this paper, we show that attention weights \textrm{DO} capture accurate word alignments and propose two novel word alignment induction methods \textproc{Shift-Att} and \textproc{Shift-AET}. The main idea is to induce alignments at the step when the to-be-aligned target token is the decoder input rather than the decoder output as in previous work. \textproc{Shift-Att} is an interpretation method that induces alignments from the attention weights of Transformer and does not require parameter update or architecture change. \textproc{Shift-AET} extracts alignments from an additional alignment module which is tightly integrated into Transformer and trained in isolation with supervision from symmetrized \textproc{Shift-Att} alignments. Experiments on three publicly available datasets demonstrate that both methods perform better than their corresponding neural baselines and \textproc{Shift-AET} significantly outperforms \textproc{GIZA++} by 1.4-4.8 AER points.\footnote{Code can be found at \url{https://github.com/sufe-nlp/transformer-alignment}.}  
\end{abstract}

\section{Introduction}\label{sec:intro}
\begin{figure}[!t]
    \centering
    \includegraphics[width=.5\textwidth]{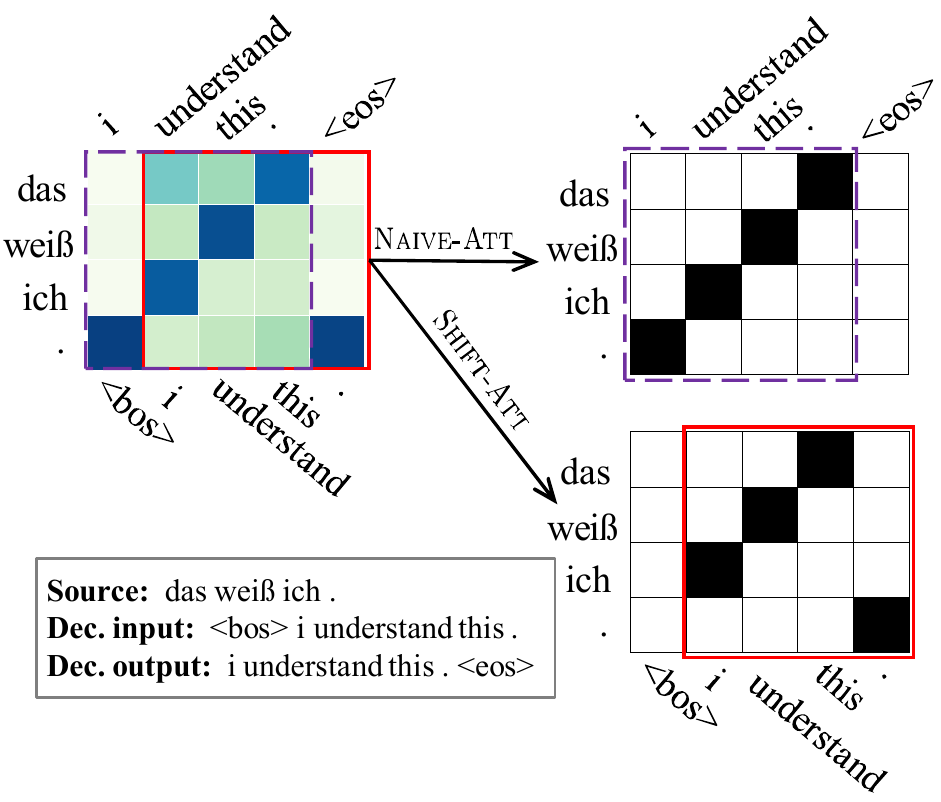}
    \caption{An example to compare our method \textproc{Shift-Att} and the baseline \textproc{Naive-Att}. The left is an attention map from the third decoder layer of the vanilla Transformer and the right are the induced alignments. \textproc{Shift-Att} induces alignments for target word $y_i$ at decoding step $i+1$ when $y_i$ is the decoder input, while \textproc{Naive-Att} at step $i$ when $y_i$ is the decoder output.} 
    \label{fig:sys}
\end{figure}
The task of word alignment is to find lexicon translation equivalents from parallel corpus~\cite{Brown93smt}. It is one of the fundamental tasks in natural language processing (NLP) and is widely studied by the community \cite{dyer-etal-2013-simple,Brown93smt,liu2015contrastive}. Word alignments are useful in many scenarios, such as error analysis~\cite{ding-etal-2017-visualizing,li-etal-2019-word}, the introduction of coverage and fertility models~\cite{tu-etal-2016-modeling}, inserting external constraints in interactive machine translation~\cite{hasler-etal-2018-neural,chenlexical} and providing guidance for human translators in computer-aided translation~\cite{dagan-etal-1993-robust}. %\cite{liu2015contrastive,ding-etal-2017-visualizing,}

Word alignment is part of the pipeline in statistical machine translation~\cite[SMT]{koehn-etal-2003-statistical}, but is not necessarily needed for neural machine translation~\cite[NMT]{Bahdanau2014NeuralMT}. The attention mechanism in NMT does not functionally play the role of word alignments between the source and the target, at least not in the same way as its analog in SMT. It is hard to interpret the attention activations and extract meaningful word alignments especially from Transformer~\cite{garg-etal-2019-jointly}. As a result, the most widely used word alignment tools are still external statistical models such as \textproc{Fast-Align} \cite{dyer-etal-2013-simple} and \textproc{GIZA++} \cite{Brown93smt,gizapp}. 

Recently, there is a resurgence of interest in the community to study word alignments for the Transformer~\cite{ding-etal-2019-saliency,li-etal-2019-word}. One simple solution is \textproc{Naive-Att}, which induces word alignments from the attention weights between the encoder and decoder. The next target word is aligned with the source word that has the maximum attention weight, as shown in Fig.~\ref{fig:sys}. However, such schedule only captures noisy word alignments~\cite{ding-etal-2019-saliency,garg-etal-2019-jointly}. One of the major problems is that it induces alignment before observing the to-be-aligned target token~\cite{peter2017generating,ding-etal-2019-saliency}. Suppose for the same source sentence, there are two alternative translations that diverge at decoding step $i$, generating $y_i$ and $y'_i$ which respectively correspond to different source words. Presumably, the source word that is aligned to $y_i$ and $y'_i$ should change correspondingly. However, this is not possible under the above method, because the alignment scores are computed before prediction of $y_i$ or $y'_i$.  %The attention weights are averaged across all heads from the penultimate decoder layer. \cite{alkhouli-etal-2018-alignment,li-etal-2019-word,}

To alleviate this problem, some researchers modify the transformer architecture by adding alignment modules that predict the to-be-aligned target token~\cite{zenkel2019adding,zenkel2020end} or modify the training loss by designing an alignment loss computed with full target sentence~\cite{garg-etal-2019-jointly,zenkel2020end}. Others argue that using only attention weights is insufficient for generating clean word alignment and propose to induce alignments with feature importance measures, such as leave-one-out measures~\cite{li-etal-2019-word} and gradient-based measures~\cite{ding-etal-2019-saliency}. However, all previous work induces alignment for target word $y_i$ at step $i$, when $y_i$ is the decoder output.

In this work, we propose to induce alignment for target word $y_i$ at step $i+1$ rather than at step $i$ as in previous work. The motivation behind this is that the hidden states in step $i+1$ are computed taking word $y_i$ as the input, thus they can incorporate the information of the to-be-aligned target token $y_i$ easily. Following this idea, we present \textproc{Shift-Att} and \textproc{Shift-AET}, two simple yet effective methods for word alignment induction. Our contributions are threefold:

\noindent $\bullet$ We introduce \textproc{Shift-Att} (see Fig. \ref{fig:sys}), a pure interpretation method to induce alignments from attention weights of vanilla Transformer. \textproc{Shift-Att} is able to reduce the Alignment Error Rate (AER) by 7.0-10.2 points over \textproc{Naive-Att} and 5.5-7.9 points over \textproc{Fast-Align} on three publicly available datasets, demonstrating that if the correct decoding step and layer are chosen, attention weights in vanilla Transformer are \textbf{sufficient} for generating accurate word alignment interpretation.

\noindent $\bullet$ We further propose \textproc{Shift-AET }, which extracts alignments from an additional alignment module. The module is tightly integrated into vanilla Transformer and trained with supervision from symmetrized \textproc{Shift-Att} alignments. \textproc{Shift-AET} does not affect the translation accuracy and significantly outperforms \textproc{GIZA++} by 1.4-4.8 AER points in our experiments.

\noindent $\bullet$ We compare our methods with \textproc{Naive-Att} on dictionary-guided decoding~\cite{alkhouli-etal-2018-alignment}, an alignment-related downstream task. Both methods consistently outperform \textproc{Naive-Att}, demonstrating the effectiveness of our methods in such alignment-related NLP tasks.

\section{Background}
\subsection{Neural Machine Translation}
Let $\mathbf{x}=\{x_1,...,x_{|\mathbf{x}|}\}$ and $\mathbf{y}=\{y_1,...,y_{|\mathbf{y}|}\}$ be source and target sentences. Neural machine translation models the target sentence given the source sentence as $p(\mathbf{y}|\mathbf{x}; \bm{\theta})$:
\begin{eqnarray}
    \label{eq:mle}
 p(\mathbf{y}|\mathbf{x}; \bm{\theta})= \prod_{t=1}^{|\mathbf{y}|+1} p(y_t|y_{0:t-1},\mathbf{x}; \bm{\theta}),
\end{eqnarray}
where $y_0=\langle \mathrm{bos}\rangle$ and $y_{|\mathbf{y}|+1}=\langle \mathrm{eos}\rangle$ represent the beginning and end of the target sentence respectively, and $\bm{\theta}$ is a set of model parameters. 

In this paper, we use Transformer~\cite{Vaswani2017AttentionIA} to implement the NMT model. Transformer is an encoder-decoder model that only relies on attention. Each decoder layer attends to the encoder output with multi-head attention. We refer to the original paper~\cite{Vaswani2017AttentionIA} for more model details. %, which consists of $N$ attention heads running in parallel. 

\subsection{Alignment by Attention} \label{sec:aba}
The encoder output from the last encoder layer is denoted as $\mathbf{h}=\{h_1,...,h_{|\mathbf{x}|}\}$, and the hidden states at decoder layer $l$ as $\mathbf{z}=\{z_1^l,...,z_{|\mathbf{y}|+1}^l\}$. For decoder layer $l$, we define the head averaged encoder-decoder attention weights as $\bm{W}^l \in \mathbb{R}^{(|\mathbf{y}|+1) \times |\mathbf{x}|}$, in which the element $W_{i,j}^l$ measures the relevance between decoder hidden state $z_i^l$ and encoder output $h_j$. For simplicity, below we use the term ``attention weights'' to denote the head averaged encoder-decoder attention weights.  

Given a trained Transformer model, word alignments can be extracted from the attention weights. More specifically, we denote the alignment score matrix as $\bm{S} \in \mathbb{R}^{{{|\mathbf{y}|}\times  |\mathbf{x}|}}$, in which the element $S_{i,j}$ is the alignment score of target word $y_i$ and source word $x_j$. Then we compute $\bm{S}$ with:
\begin{equation}
S_{i,j}=W_{i,j}^l\quad(1\leq i\leq |\mathbf{y}|\mathrm{, } 1\leq j\leq |\mathbf{x}|)
\label{eq:score_att}
\end{equation}
and extract word alignments $\bm{A}$ with maximum a posterior strategy following~\citet{garg-etal-2019-jointly}:
\begin{eqnarray}
    A_{ij} =  \left\{
    \begin{array}{cl} 
    1 & \text{if  } j=\argmax_{j'} S_{i,j'}  \\[0.05cm]
    0  & \text{otherwise}
    \end{array}
    \right. , 
    \label{eq:align}
\end{eqnarray}
where $A_{ij}=1$ indicates $y_i$ is aligned to $x_j$. We call this approach \textproc{Naive-Att}. \citet{garg-etal-2019-jointly} show that attention weights from the penultimate layer, i.e., $l=L-1$, can induce the best alignments. %, ,  under their setting

Although simple to implement, this method fails to obtain satisfactory word alignments~\cite{ding-etal-2019-saliency,garg-etal-2019-jointly}. First of all, instead of the relevance between $y_i$ and $x_j$, $W_{i,j}^l$ measures the relevance between decoder hidden state $z_i^l$ and encoder output $h_j$. Considering that the decoder input is $y_{i-1}$ and the output is $y_i$ at step $i$, $z_i^l$ may better represent $y_{i-1}$ instead of $y_i$, especially for bottom layers. Second, since $W_{i,j}^l$ is computed before observing $y_i$, it becomes difficult for it to induce the aligned source token for the target token $y_i$, as discussed in Section~\ref{sec:intro}.

As a result, it is necessary to develop novel methods for alignment induction. This method should be able to (\romannumeral1) take into account the relationship of $z_i^l$, $y_i$ and $y_{i-1}$, and (\romannumeral2) adapt the alignment induction with the to-be-aligned target token.

\section{Method}
In this section, we propose two novel alignment induction methods \textproc{Shift-Att} and \textproc{Shift-AET}. Both methods adapt the alignment induction with the to-be-aligned target token by computing alignment scores at the step when the target token is the decoder input. 

\subsection{\textproc{Shift-Att}: Alignment from Vanilla Transformer}
\paragraph{Alignment Induction}\label{sec:ttr}
\textproc{Naive-Att}~\cite{garg-etal-2019-jointly} induces alignment for target token $y_i$ at step $i$ when $y_i$ is the decoder output and defines the alignment score matrix with Eq.~\ref{eq:score_att}. They find the best layer $l$ to extract alignments by evaluating the AER of all layers on the test set.

We instead propose to induce alignment for target token $y_i$ at step $i+1$ when $y_i$ is the decoder input. We define the alignment score matrix $\bm{S}$ as:
\begin{equation}
S_{i,j}=W_{i+1,j}^l \quad (1\leq i\leq|\mathbf{y}|\mathrm{, }1\leq j\leq |\mathbf{x}|).
\end{equation}
This is because $W_{i+1,j}^l$ measures the relevance between $z_{i+1}^l$ and $h_j$, and we use $z_{i+1}^l$ and $h_j$ to represent $y_i$ and $x_j$ respectively. With the alignment score matrix $\bm{S}$, we can extract word alignments $\bm{A}$ using Eq.~\ref{eq:align}. We call this method \textproc{Shift-Att}. Fig. \ref{fig:sys} shows an alignment induction example to compare \textproc{Naive-Att} and \textproc{Shift-Att}. 

\textproc{Shift-Att} uses $z_{i+1}^l$ to represent the to-be-aligned target token $y_i$ while \textproc{Naive-Att} uses $z_i^l$. We argue using $z_{i+1}^l$ is better. First, at bottom layers, we hypothesize that $z_{i+1}^l$ could better represent the decoder input $y_i$ than output $y_{i+1}$. Therefore we can use $z_{i+1}^l$ with small $l$ to represent $y_i$. Second, $z_{i+1}^l$ is computed after observing $y_i$, indicating that \textproc{Shift-Att} is able to adapt the alignment induction with the to-be-aligned target token. 

Our proposed method involves inducing alignments from source-to-target and target-to-source vanilla Transformer models. Following \citet{zenkel2019adding}, we merge bidirectional alignments using the grow diagonal heuristic~\cite{koehn2005edinburgh}.

\paragraph{Layer Selection Criterion}\label{sec:lscc}
To select the best layer $l_b$ to induce alignments, we propose a surrogate layer selection criterion without manually labelled word alignments. Experiments show that this criterion correlates well with the AER metric.

Given parallel sentence pairs $\langle \mathbf{x}, \mathbf{y}\rangle$, we train a source-to-target model $\bm{\theta}_{\mathbf{x} \rightarrow \mathbf{y}} $ and a target-to-source model $\bm{\theta}_{\mathbf{y} \rightarrow \mathbf{x}} $. We assume that the word alignments extracted from these two models should agree with each other~\cite{cheng2016agreement}. Therefore, we evaluate the quality of the alignments by computing the AER score on the validation set with the source-to-target alignments as the hypothesis and the target-to-source alignments as the reference. For each model, we can obtain $L$ word alignments from $L$ different layers. In total, we obtain $L\times L$ AER scores. We select the one with the lowest AER score, and its corresponding layers of the source-to-target and target-to-source models are the layers we will use to extract alignments at test time:
\begin{equation}
% \resizebox{.43 \textwidth}{!}{$
l_{\text{b},\mathbf{x} \rightarrow \mathbf{y}},l_{\text{b}, \mathbf{y} \rightarrow \mathbf{x}}\\
=\argmin_{i,j}{\text{AER}(\bm{A}_{\mathbf{x} \rightarrow \mathbf{y}}^i, \bm{A}_{\mathbf{y} \rightarrow \mathbf{x}}^j)}. \nonumber
% $}
\label{eq:sel}
\end{equation}

\subsection{\textproc{Shift-AET}: Alignment from Alignment-Enhanced Transformer}
\begin{figure}[!t]
    \centering
    \includegraphics[width=.5\textwidth]{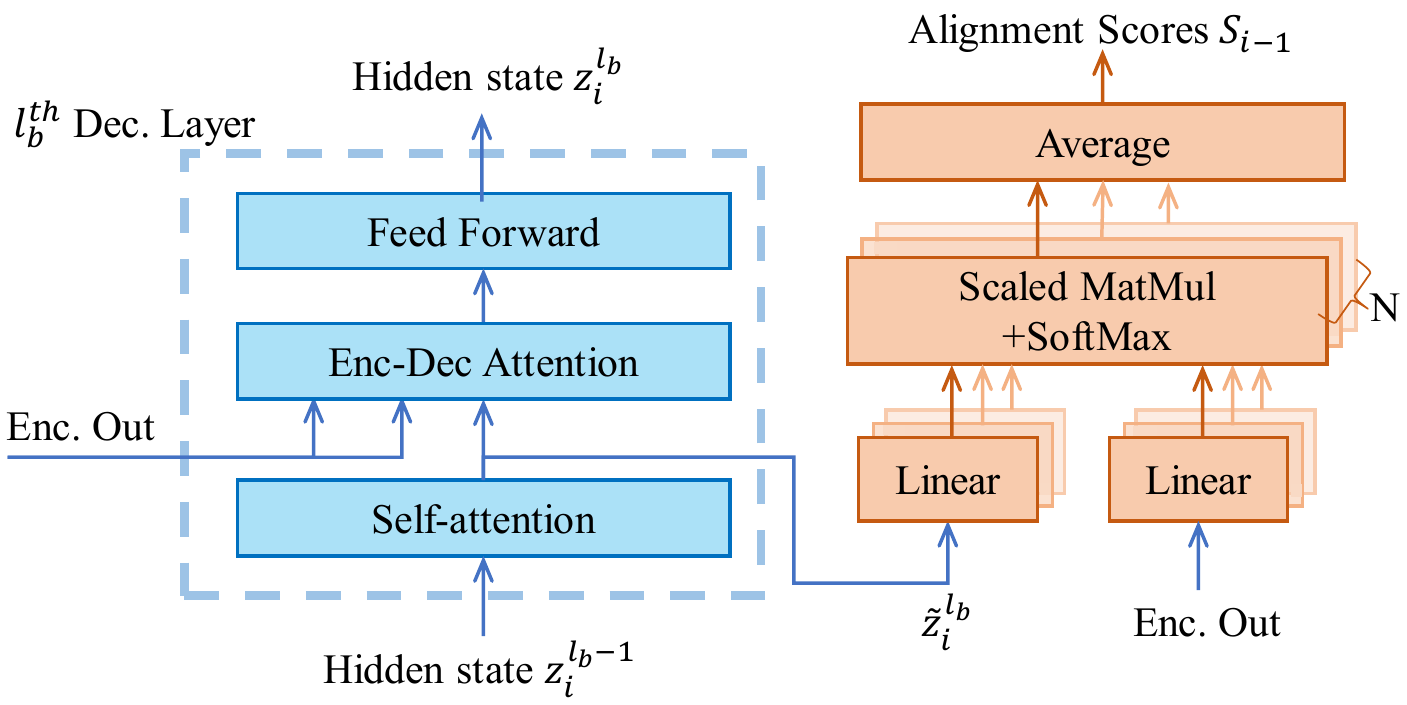}
    \caption{Illustration of the alignment module at decoding step $i$. The decoder input token is $y_{i-1}$, while the output token is $y_i$. The alignment module predicts $S_{i-1}$, the alignment scores corresponding to the input target token $y_{i-1}$. During the alignment module training process, parameters of the blue blocks are frozen, and only parameters of the orange blocks are updated.} 
    \label{fig:align_sys}
\end{figure}
To further improve the alignment accuracy, we propose \textproc{Shift-AET}, a word alignment induction method that extracts alignments from Alignment-Enhanced Transformer (AET). AET extends the Transformer architecture with a separate alignment module, which observes the hidden states of the underlying Transformer at each step and predicts the alignment scores for the current decoder \textbf{input}. Note that this module is a plug and play component and it neither makes any change to the underlying NMT model nor influences the translation quality.  

Fig. \ref{fig:align_sys} illustrates the alignment module of AET at decoding step $i$. We add the alignment module only at layer $l_b$, the best layer to extract alignments with \textproc{Shift-Att}. The alignment module performs multi-head attention similar to the encoder-decoder attention sublayer. It takes the encoder outputs $\mathbf{h}=\{h_1,...,h_{|\mathbf{x}|}\}$ and the current decoder hidden state $\tilde{z}_i^{l_b}$ inside layer $l_b$ as input and outputs $S_{i-1}$, the alignment score corresponding to target word $y_{i-1}$:
\begin{equation}
    S_{i-1}
    =\frac{1}{N} \sum_n \mathrm{softmax}(\frac{(\mathbf{h}\bm{G}_n^K) (\tilde{z}_i^{l_b}\bm{G}_n^Q) ^{\top}}{\sqrt{d_k}}),
\end{equation}
where  $\bm{G}_n^K\mathrm{, }\bm{G}_n^Q \in \mathbb{R}^{d_{
\mathrm{model}} \times d_k}$ are the key and query projection matrices for the $n$-th head, $N$ is the number of attention heads and $d_k=d_{
\mathrm{model}}/N$. Since we only care about the attention weights, the value-related parameters and computation are omitted in this module. 

To train the alignment module, we use the symmetrized \textproc{Shift-Att} alignments extracted from vanilla Transformer models as labels. Specifically, while the underlying Transformer is pretrained and fixed (Fig.~\ref{fig:align_sys}), we train the alignment module with the loss function following~\citet{garg-etal-2019-jointly}:
\begin{eqnarray}
\mathcal{L}_{a} = -\frac{1}{|\mathbf{y}|}\sum_{i = 1}^{|\mathbf{y}|}\sum_{j = 1}^{|\mathbf{x}|} \big(\hat{A}^{p}_{i,j} \odot \log S_{i,j}),
\end{eqnarray}
where $\bm{S}=\{S_1\text{;}...\text{;}S_{|\mathbf{y}|}\}$ is the alignment score matrix predicted by the alignment module, and $\bm{\hat{A}}^{p}$ denotes the normalized reference symmetrized \textproc{Shift-Att} alignments.\footnote{We simply normalize rows corresponding to target tokens that are aligned to at least one source token of $\bm{\hat{A}}$.} In this way, we transfer the alignment knowledge implicitly learned in two vanilla Transformer models  $\bm{\theta}_{\mathbf{x} \rightarrow \mathbf{y}} $  and  $\bm{\theta}_{\mathbf{y} \rightarrow \mathbf{x}} $ into the alignment module of a single AET model.  %the alignment matrix

Once the alignment module is trained, we extract alignment scores $\bm{S}$  from it given a parallel sentence pair and induce alignments $\bm{A}$ using Eq.~\ref{eq:align}.

\begin{table}[!t]
	\centering
	\small
	\begin{tabular}{l | r r r}
    	\hline
		Dataset & Train & Validation & Test \\ \hline
        de-en & 1.9M & 994 & 508  \\ 
        fr-en & 1.1M & 1,000 & 447  \\
        ro-en & 0.5M & 999 & 248 \\ 
         \hline
	\end{tabular}
	\caption{Number of sentences in each dataset.}\label{table:stat}
\end{table}

\section{Experiments}
\begin{table*}[!t]
\resizebox{2.0\columnwidth}{!}{
	\centering
	\begin{tabular}{ l | r r | r r r | r r r | r r r }
    	\hline
		  \multirow{2}{*}{Method} & \multirow{ 2}{*}{Inter.} & \multirow{ 2}{*}{Fullc} & \multicolumn{3}{c|}{de-en} & \multicolumn{3}{c|}{fr-en} & \multicolumn{3}{c}{ro-en} \\ 
		  & & &  de$\rightarrow$en & en$\rightarrow$de & bidir & fr$\rightarrow$en & en$\rightarrow$fr & bidir & ro$\rightarrow$en & en$\rightarrow$ro & bidir \\ \hline \hline
		 \multicolumn{12}{c}{\textit{Statistical Methods}} \\
		 \hline
         \textproc{Fast-Align}~\cite{dyer-etal-2013-simple}  &  - & Y & 28.5 & 30.4 & 25.7 & 16.3 & 17.1 & 12.1 & 33.6 & 36.8 & 31.8  \\ 
         \textproc{GIZA++}~\cite{Brown93smt}  &  - & Y & 18.8 & 19.6 & 17.8 & 7.1 & 7.2 & 6.1 & 27.4 & 28.7 & 26.0 \\  \hline
		 \multicolumn{12}{c}{\textit{Neural Methods}} \\
		 \hline
         \textproc{Naive-Att}~\cite{garg-etal-2019-jointly} & Y & N & 33.3 & 36.5 & 28.1 & 27.5 & 23.6 & 16.0 & 33.6 & 35.1 & 30.9  \\ 
         \textproc{Naive-Att-LA}~\cite{garg-etal-2019-jointly} & Y & N & 40.9 & 50.8 & 39.8 & 32.4 & 29.8 & 21.2 & 37.5 & 35.5 & 32.7  \\ 
         \textproc{Shift-Att-LA} & Y & N & 54.7 & 46.2 & 45.5 & 60.5 & 46.9 & 55.1 & 66.1 & 60.4 & 65.3  \\ 
         \textproc{SmoothGrad}~\cite{li-etal-2016-visualizing} &  Y & N & 36.4 & 45.8 & 30.3 & 25.5 & 27.0 & 15.6 & 41.3 & 39.9 & 33.7 \\ 
         \textproc{SD-SmoothGrad}~\cite{ding-etal-2019-saliency}  &  Y & N & 36.4 & 43.0 & 29.0 & 25.9 & 29.7  & 15.3 & 41.2 & 41.4 & 32.7  \\ 
         \textproc{PD}~\cite{li-etal-2019-word}  & Y & N & 38.1 & 44.8 & 34.4 & 32.4 & 31.1 & 23.1 & 40.2 & 40.8 & 35.6  \\

        %  \textit{naive-loo} & -  & N & 36.6 & 42.3 & 33.0 & 29.5 & 29.2 & 20.4 & 36.5 & 36.9 & 32.7  \\ 

        %  \textit{\textproc{Naive-Att}-mtl}~\shortcite{garg-etal-2019-jointly} & loss & N & 23.5 & 25.3 & 21.3 & 16.3 & 15.2 & 11.1 &  31.0 & 28.3 & 28.3  \\
        \textproc{AddSGD}~\cite{zenkel2019adding} & N & N & 26.6 & 30.4 & 21.2 & 20.5 & 23.8 & 10.0 & 32.3 & 34.8 & 27.6   \\ 
        \textproc{Mtl-Fullc}~\cite{garg-etal-2019-jointly} & N & Y & - & - & 20.2 & - & - & 7.7  & - & - & 26.0  \\ \hline
        \multicolumn{12}{c}{\textit{Statistical + Neural Methods}} \\
		 \hline
        \textproc{Mtl-Fullc-GZ}~\cite{garg-etal-2019-jointly} & N & Y & - & - & 16.0 & - & - & \textbf{4.6}  & - & - & 23.1 \\ \hline
		 \multicolumn{12}{c}{\textit{Our Neural Methods}} \\
		 \hline
        %  \textit{align-loo}  & - & N & 31.5 & 34.5 & 24.4 & 25.4 & 27.3 & 13.9 & 34.2 & 34.6 & 28.7 \\ 
        \textproc{Shift-Att}  & Y  & N & 20.9 & 25.7 & \underline{17.9} & 17.1 & 16.1 & \underline{6.6} & 27.4 & 26.0 & \underline{23.9} \\ 
        \textproc{Shift-AET} & N & N & 15.8 & 19.2 & \textbf{15.4} & 9.9 & 10.5 & 4.7  & 22.7 & 23.6 & \textbf{21.2}  \\ \hline
	\end{tabular}}
	\caption{AER on the test set with different alignment methods. \emph{bidir} are symmetrized alignment results. The column Inter. represents whether the method is an interpretation method that can extract alignments from a pretrained vanilla Transformer model. The column Fullc denotes whether full target sentence is used to extract alignments at test time. The lower AER, the better. We mark best symmetrized interpretation results of vanilla Transformer with underlines, and best symmetrized results among all with boldface. }\label{table:aer}
\end{table*}

\subsection{Settings}\label{sec:data}
\paragraph{Dataset}
We follow previous work~\cite{zenkel2019adding,zenkel2020end} in data setup and conduct experiments on publicly available datasets for German-English (de-en)\footnote{\url{https://www-i6.informatik.rwth-aachen.de/goldAlignment/}}, Romanian-English (ro-en) and French-English (fr-en)\footnote{\url{http://web.eecs.umich.edu/˜mihalcea/wpt/index.html}}. Since no validation set is provided, we follow \citet{ding-etal-2019-saliency} to set the last 1,000 sentences of the training data before preprocessing as validation set. We learn a joint source and target Byte-Pair-Encoding~\cite{Sennrich2016NeuralMT} with 10k merge operations. Table~\ref{table:stat} shows the detailed data statistics. 

\paragraph{NMT Systems}
We implement the Transformer with fairseq-py\footnote{\url{https://github.com/pytorch/fairseq}} and use the \texttt{transformer\_iwslt\_de\_en} model configuration following~\citet{ding-etal-2019-saliency}. We train the models with a batch size of 36K tokens and set the maximum updates as 50K and 10K for Transformer and AET respectively. The last checkpoint of AET is used for evaluation. All models are trained in both translation directions and symmetrized with \textit{grow-diag}~\cite{koehn2005edinburgh} using the script from \citet{zenkel2019adding}.\footnote{\url{https://github.com/lilt/alignment-scripts}} 

\paragraph{Evaluation}
We evaluate the alignment quality of our methods with Alignment Error Rate \cite[AER]{och2000improved}. Since word alignments are useful for many downstream tasks as discussed in Section \ref{sec:intro}, we also evaluate our methods on dictionary-guided decoding, a downstream task of alignment induction, with the metric BLEU~\cite{Papineni2002BleuAM}. More details are in Section \ref{sec:dgd}. 

\paragraph{Baselines} 
We compare our methods with two statistical baselines \textproc{Fast-Align} and \textproc{GIZA++} and nine other baselines:
% \begin{itemize} [noitemsep,leftmargin=*] %topsep=0pt,parsep=0pt,partopsep=0pt,

\noindent $\bullet$ \textproc{Naive-Att}~\cite{garg-etal-2019-jointly}: the approach we discuss in Section \ref{sec:aba}, which induces alignments from the attention weights of the penultimate layer of the Transformer.

\noindent $\bullet$ \textproc{Naive-Att-LA}~\cite{garg-etal-2019-jointly}: the \textproc{Naive-Att} method without layer selection. It induces alignments from attention weights averaged across all layers. 

\noindent $\bullet$ \textproc{Shift-Att-LA}: \textproc{Shift-Att} method without layer selection. It induces alignments from attention weights averaged across all layers. 

\noindent $\bullet$ \textproc{SmoothGrad}~\cite{li-etal-2016-visualizing}: the method that induces alignments from word saliency, which is computed by averaging the gradient-based saliency scores with multiple noisy sentence pairs as input. 

\noindent $\bullet$ \textproc{SD-SmoothGrad}~\cite{ding-etal-2019-saliency}: an improved version of \textproc{SmoothGrad}, which defines saliency on one-hot input vector instead of word embedding.
 
\noindent $\bullet$ \textproc{PD}~\cite{li-etal-2019-word}: the method that computes the alignment scores from Transformer by iteratively masking each source token and measuring the prediction difference. 

\noindent $\bullet$ \textproc{AddSGD}~\cite{zenkel2019adding}: the method that explicitly adds an extra attention layer on top of Transformer and directly optimizes its activations towards predicting the to-be-aligned target token. 

\noindent $\bullet$ \textproc{Mtl-Fullc}~\cite{garg-etal-2019-jointly}: the method that trains a single model in a multi-task learning framework to both predict the target sentence and the alignment. When predicting the alignment, the model observes full target sentence and uses symmetrized \textproc{Naive-Att} alignments as labels.

\noindent $\bullet$ \textproc{Mtl-Fullc-GZ}~\cite{garg-etal-2019-jointly}: the same method as \textproc{Mtl-Fullc} except using symmetrized \textproc{GIZA++} alignments as labels. It is a statistical and neural method as it relies on \textproc{GIZA++} alignments.

Among these nine baselines and our proposed methods, \textproc{SmoothGrad}, \textproc{SD-SmoothGrad} and \textproc{PD} induce alignments using feature importance measures, while the others from some form of attention weights. Note that the computation cost of methods with feature importance measures is much higher than those with attention weights.\footnote{For each sentence pair, \textproc{PD} forwards once with $|\mathbf{x}|+1$ masked sentence pairs as the input, while \textproc{SmoothGrad} and \textit{\textproc{SD-SmoothGrad}} forward and backward once with $m$ ($m=30$ in \citet{ding-etal-2019-saliency}) noisy sentence pairs as the input. In contrast, attention weights based methods forward once with one sentence pair as the input.} 

\subsection{Alignment Results}
\paragraph{Comparison with Baselines}
Table \ref{table:aer} compares our methods with all the baselines. First, \textproc{Shift-Att}, a pure interpretation method for the vanilla Transformer, significantly outperforms \textproc{Fast-Align} and all neural baselines, and performs comparable with \textproc{GIZA++}. For example, it outperforms \textit{\textproc{SD-SmoothGrad}}, the state-of-the-art method with feature importance measures to extract alignments from vanilla Transformer, by $8.7$-$11.1$ AER points across different language pairs. The success of \textproc{Shift-Att} demonstrates that vanilla Transformer has captured alignment information in an implicit way, which could be revealed from the attention weights if the correct decoding step and layer are chosen to induce alignments.

Second, the method \textproc{Shift-AET} achieves new state-of-the-art, significantly outperforming all baselines. It improves over \textproc{GIZA++} by $1.4$-$4.8$ AER across different language pairs, demonstrating that it is possible to build a neural aligner better than \textproc{GIZA++} without using any alignments generated from statistical aligners to bootstrap training. We also find \textproc{Shift-AET} performs either marginally better (de-en and ro-en) or on-par (fr-en) when comparing with \textproc{MTL-Fullc-GZ}, a method that uses \textproc{GIZA++} alignments to bootstrap training. We evaluate the model sizes: the number of parameters in vanilla Transformer and AET are 36.8M and 37.3M respectively, and find that AET only introduces 1.4\% additional parameters to the vanilla Transformer. In summary, by supervising the alignment module with symmetrized \textproc{Shift-Att} alignments, \textproc{Shift-AET} improves over \textproc{Shift-Att} and \textproc{GIZA++} with negligible parameter increase and without influencing the translation quality.

\paragraph{Comparison with \citet{zenkel2020end}}
\begin{table}[!t]
	\centering
	\small
% 	\begin{subtable}{.5\textwidth}
	\begin{tabular}{l | l l l }
    	\hline
		Method & de-en & fr-en & ro-en \\ \hline
        \textproc{BAO-Guided} & 16.3 & 5.0 & 23.4  \\ 
        \textproc{Shift-AET} & \textbf{15.4} & \textbf{4.7} & \textbf{21.2}  \\ \hline
	\end{tabular}
% 	\end{subtable}
	\caption{Comparison of our method \textproc{Shift-AET} with \textproc{BAO-Guided}~\cite{zenkel2020end}. We report the symmetrized AER on the test set.}\label{table:e2e}
\end{table}
Concurrent with our work, \citet{zenkel2020end} propose a neural aligner that can outperform \textproc{GIZA++}. Table \ref{table:e2e} compares the performance of \textproc{Shift-AET} and the best method \textproc{BAO-Guided} (Birdir. Att. Opt. + Guided) in \citet{zenkel2020end}. We observe that \textproc{Shift-AET} performs better than \textproc{BAO-Guided} in terms of alignment accuracy.

\textproc{Shift-AET} is also much simpler than \textproc{BAO-Guided}. The training of \textproc{BAO-Guided} includes three stages: (\romannumeral1) train vanilla Transformer in source-to-target and target-to-source directions; (\romannumeral2) train the alignment layer and extract alignments on the training set with bidirectional attention optimization. This alignment extraction process is computational costly since bidirectional attention optimization fine-tunes the model parameters separately for each sentence pair in the training set; (\romannumeral3) re-train the alignment layer with the extracted alignments as the guidance. In contrast, \textproc{Shift-AET} can be trained much faster in two stages and does not involve 
bidirectional attention optimization. 

Similar with \textproc{Mtl-Fullc}~\cite{garg-etal-2019-jointly}, \textproc{BAO-Guided} adapts the alignment induction with the to-be-aligned target token by requiring full target sentence as the input. Therefore, \textproc{BAO-Guided} is not applicable in cases where alignments are incrementally computed during the decoding process, e.g., dictionary-guided decoding~\cite{alkhouli-etal-2018-alignment}. In contrast, \textproc{Shift-AET} performs quite well on such cases (Section~\ref{sec:dgd}). Therefore, considering the alignment performance, computation cost and applicable scope, we believe \textproc{Shift-AET} is more appropriate than \textproc{BAO-Guided} for the task of alignment induction.

\begin{table}[!t]
	\centering
	\small
	\begin{tabular}{l | r r r}
    	\hline
		Direction & zh$\rightarrow$en & en$\rightarrow$zh & bidir \\ \hline
        \textproc{GIZA++} & 19.6 & 23.3 & 18.5  \\ 
        \textproc{Naive-Att} & 36.9 & 40.3 & 28.9  \\ \hline
        \textproc{Shift-Att} & 28.1 & 27.3 & 20.2 \\ 
        \textproc{Shift-AET} & 20.1 & 22.0 & \textbf{17.2} \\ 
         \hline
	\end{tabular}
	\caption{AER on the test set of zh-en. \emph{bidir} are symmetrized alignment results.}\label{table:distant}
\end{table}

\paragraph{Performance on Distant Language Pair} To further demonstrate the superiority of our methods on distant language pairs, we also evaluate our methods on Chinese-English (zh-en). We use NIST corpora\footnote{The corpora include LDC2002E18, LDC2003E07, LDC2003E14, LDC2004T07, LDC2004T08 and LDC2005T06} as the training set and v1-tstset released by TsinghuaAligner~\cite{liu2015contrastive} as the test set. The test set includes 450 parallel sentence pairs with manually labelled word alignments.\footnote{TsinghuaAligner labels the word alignments based on segmented Chinese sentences and does not provide the segmentation model. Therefore, we convert the manually labelled word alignments to our segmented Chinese sentences for evaluation.} We use \textit{jieba}\footnote{\url{https://github.com/fxsjy/jieba}} for Chinese text segmentation and follow the settings in Section~\ref{sec:data} for data pre-processing and model training. The results are shown in Table~\ref{table:distant}. It presents that both \textproc{Shift-Att} and \textproc{Shift-AET} outperform \textproc{Naive-Att} to a large margin. When comparing the symmetrized alignment performance with \textproc{GIZA++}, \textproc{Shift-AET} performs better, while \textproc{Shift-Att} is worse. The experimental results are roughly consistent with the observations on other language pairs, demonstrating the effectiveness of our methods even for distant language pairs.

\subsection{Downstream Task Results}\label{sec:dgd}
\begin{table}[!t]
% \resizebox{1.0\columnwidth}{!}{
	\centering
	\small
	\begin{tabular}{l|l l l }
    	\hline
		Task & \textproc{Naive-Att} & \textproc{Shift-Att} & \textproc{Shift-AET} \\ \hline
        de$\rightarrow$en & 33.7 & 34.3$^*$ & \textbf{34.8}$^*$  \\ 
        en$\rightarrow$de & 26.5 & 26.8 & \textbf{28.0}$^*$ \\  
         \hline
	\end{tabular} %}
	\caption{Comparison of dictionary-guided decoding with different alignment methods. We report BLEU scores on the test set. Without dictionary-guided decoding, we obtain
	32.3 and 24.2 BLEU on de$\rightarrow$en and en$\rightarrow$de translations respectively. “*” indicates the result is significantly better than that of \textproc{Naive-Att}  (p$<$0.05). All significance tests are measured by paired bootstrap resampling~\cite{koehn2004statistical}}\label{table:dgd}
\end{table}
In addition to AER, we compare the performance of \textproc{Naive-Att}, \textproc{Shift-Att} and \textproc{Shift-AET} on dictionary-guided machine translation~\cite{song2020alignment}, which is an alignment-based downstream task. Given source and target constraint pairs from dictionary, the NMT model is encouraged to translate with provided constraints via word alignments~\cite{alkhouli-etal-2018-alignment,hasler-etal-2018-neural,hokamp-liu-2017-lexically,song2020alignment}. More specifically, at each decoding step, the last token of the candidate translation will be revised with target constraint if it is aligned to the corresponding source constraint according to the alignment induction method. To simulate the process of looking up dictionary, we follow~\citet{hasler-etal-2018-neural} and extract the pre-specified constraints from the test set and its reference according to the golden word alignments. We exclude stop words, and sample up to 3 dictionary constraints per sentence. Each dictionary constraint includes up to 3 source tokens. 

Table~\ref{table:dgd} presents the performance with different alignment methods. Both \textproc{Shift-Att} and \textproc{Shift-AET} outperform \textproc{Naive-Att}. \textproc{Shift-AET} obtains the best translation quality, improving over \textproc{Naive-Att} by 1.1 and 1.5 BLEU scores on de$\rightarrow$en and en$\rightarrow$de translations, respectively. The results suggest the effectiveness of our methods in application to alignment-related NLP tasks.

\subsection{Analysis}
\paragraph{Layer Selection Criterion} \label{sec:lsc}
\begin{table}[!t]
	\centering
	\small
	\begin{subtable}{.5\textwidth}
	\subcaption{Validation AER for Layer Selection}
	\begin{tabular}{ c | c  c  c  c  c  c }
    	\hline 
    	\diagbox[width=5em,trim=l]{en$\rightarrow$de}{de$\rightarrow$en}  & 1 & 2 & 3 & 4 & 5 & 6 \\ \hline
    	 1 & 42.2 & 35.4 & 35.7 & 67.5 & 89.2 & 88.8 \\
    	 2 & 45.1 & 39.5 & 39.1 & 67.1 & 87.8 & 88.2 \\
    	 3 & 42.5 & 34.6 & \textbf{34.2} & 65.2 & 87.4 & 87.6 \\
    	 4 & 74.4 & 73.0 & 72.3 & 80.6 & 89.5 & 89.7 \\ 
    	 5 & 84.8 & 86.7 & 86.1 & 87.3 & 88.7 & 88.9 \\ 
    	 6 & 87.1 & 88.2 & 87.6 & 88.1 & 88.7 & 88.6 \\ \hline
	\end{tabular}
	\end{subtable}
	
	\bigskip
	\begin{subtable}{.5\textwidth}
    % \vspace{+0.2cm}
	\subcaption{Test AER for Verification}
	\begin{tabular}{l | c  c  c  c  c  c }
    	\hline
    	 layer & 1 & 2 & 3 & 4 & 5 & 6 \\ \hline
    	 de$\rightarrow$en & 31.5 & 22.7 & \textbf{20.9} & 55.7 & 80.5 & 81.5  \\
    	 en$\rightarrow$de & 27.4 & 31.3 & \textbf{25.7} & 68.5 & 83.4 & 85.1 \\ \hline
	\end{tabular}
	\end{subtable}
	\caption{Layer selection criterion verification with \textproc{Shift-Att} on de-en alignment. (a) For each cell, we induce hypothesis alignment from de$\rightarrow$en translation and reference alignment from en$\rightarrow$de translation. $l_{\text{b}}=3$ for both translation directions in this table. (b) Test AER when inducing alignments from different layers. Layer $3$ induces the best alignment for both translation directions, which verifies $l_{\text{b}}$ selected in (a).}
	 \label{table:att-sel}
\end{table}
To test whether the layer selection criterion can select the right layer to extract alignments, we first determine the best layer $l_{\text{b},\mathbf{x} \rightarrow \mathbf{y}}$ and $l_{\text{b},\mathbf{y} \rightarrow \mathbf{x}}$ based on the layer selection criterion. Then we evaluate the AER scores of alignments induced from different layers on the test set, and check whether the layers with the lowest AER score are consistent with $l_{\text{b},\mathbf{x} \rightarrow \mathbf{y}}$ and $l_{\text{b},\mathbf{y} \rightarrow \mathbf{x}}$. The experiment results shown in Table~\ref{table:att-sel} verify that the layer selection criterion is able to select the best layer to induce alignments. We also find that the best layer is always layer 3 under our setting, consistent across different language pairs. 
\paragraph{Relevance Measure Verification}
To investigate the relationship between $z_i^l$ and $y_{i-1}/y_i$, we design an experiment to probe whether $z_i^l$ contain the identity information of $y_{i-1}$ and $y_i$, following~\citet{Brunner2019OnII}. Formally, for decoder hidden state $z_i^l$, the input token is identifiable if there exists a function $g$ such that $y_{i-1}=g(z_i^l)$. We cannot prove the existence of $g$ analytically. Instead, for each layer $l$ we learn a projection function $\hat{g}_l$ to project from the hidden state space to the input token embedding space $\hat{y}_i^l=\hat{g}_l(z_i^l)$ and then search for the nearest neighbour $y_k$ within the same sentence. We say that $z_i^l$ can identify $y_{i-1}$ if $k=i-1$. Similarly, we follow the same process to identify the output token $y_i$. We report the identifiability rate defined as the percentage of correctly identified tokens. 

\begin{figure}[!t]
    \centering
    \includegraphics[width=.5\textwidth]{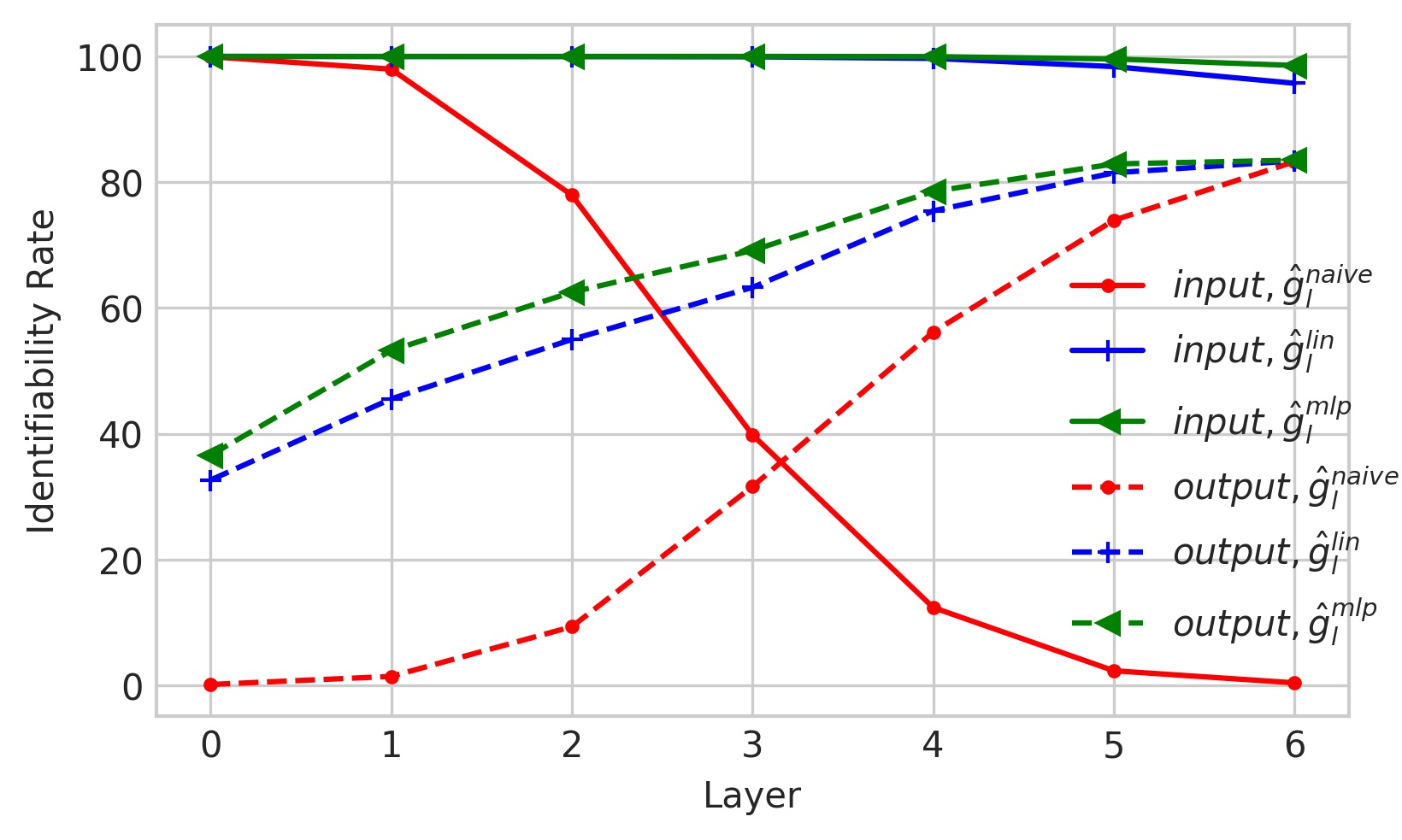}
    \caption{Identifiability rate of the input and output tokens for decoder hidden states at different layers.}
    \label{fig:iden}
\end{figure}

Fig.~\ref{fig:iden} presents the results on the validation set of de$\rightarrow$en translation. We try three projection functions: a naive baseline $\hat{g}_l^{\text{naive}}(z_i^l)=z_i^l$, a linear perceptron $\hat{g}_l^{\text{lin}}$ and a non-linear multi-layer perceptron $\hat{g}_l^{\text{mlp}}$. We observe the following points: (\romannumeral1) With trainable projection functions $\hat{g}_l^{\text{lin}}$ and $\hat{g}_l^{\text{mlp}}$, all layers can identify the input tokens, although more hidden states cannot be mapped back to their input tokens anymore in higher layers.  (\romannumeral2) Overall it is easier to identify the input token than the output token. For example, when projecting with mlp, all layers can identify more than 98\% of the input tokens. However, for the output tokens, we can only identify 83.5\% even from the best layer. Since $z_i^l$ even may not be able to identify $y_i$, this observation partially verifies that it is better to represent $y_i$ using $z_{i+1}^l$ than $z_{i}^l$. (\romannumeral3) At bottom layers, the input tokens remain identifiable and the output tokens are hard to identify, regardless of the projection function we use. This confirms our hypothesis that for small $l$, $z_i^l$ is more relevant to $y_{i-1}$ than $y_i$.

\paragraph{AER v.s. BLEU}
During training, vanilla Transformer gradually learns to align and translate. To analyze how the alignment behavior changes at different layers with checkpoints of different translation quality, we plot AER on the test set v.s. BLEU on the validation set for de$\rightarrow$en translation. We compare \textproc{Naive-Att} and \textproc{Shift-Att}, which align the decoder output token (\textit{align output}) and decoder input token (\textit{align input}) to the source tokens based on current decoder hidden state, respectively. 

\begin{figure*}[!t]
    \centering
    \includegraphics[width=\textwidth]{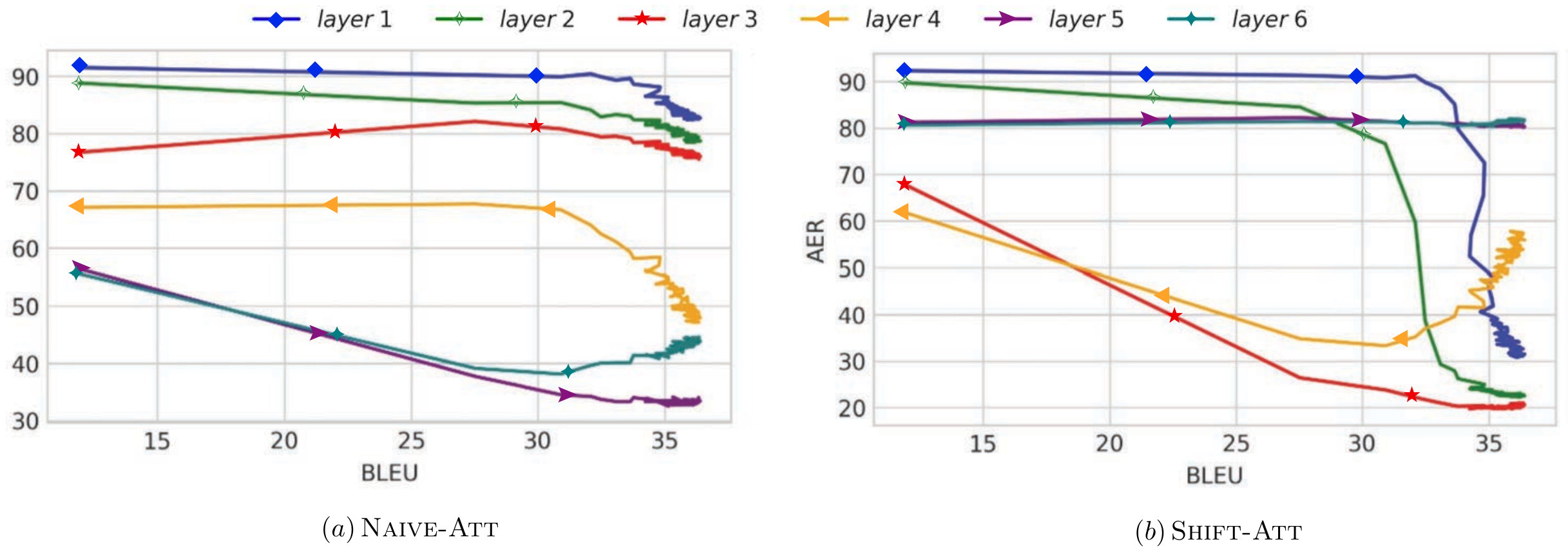}  %{figure/aer_bleu_deen.jpg}
    \caption{AER on the test set v.s. BLEU on the validation set on the de$\rightarrow$en translation, evaluated with different checkpoints.}
    \label{fig:aer_bleu}
\end{figure*}

The experiment results are shown in Fig.~\ref{fig:aer_bleu}. We observe that at the beginning of training, layers $3$ and $4$ learn to align the input token, while layers $5$ and $6$ the output token. However, with the increasing of BLEU score, layer $4$ tends to change from aligning input token to aligning output token, and layer $1$ and $2$ begin to align input token. This suggests that vanilla Transformer gradually learns to align the input token from middle layers to bottom layers. We also see that at the end of training, layer $6$'s ability to align output token decreases. We hypothesize that layer $5$ already has the ability to attend to the source tokens which are aligned to the output token, therefore attention weights in layer $6$ may capture other information needed for translation. Finally, for checkpoints with the highest BLEU score, layer $5$ aligns the output token best and layer $3$ aligns the input token best. 
\paragraph{Alignment Example}
\begin{figure*}[!ht]
    \centering
    \includegraphics[width=\textwidth]{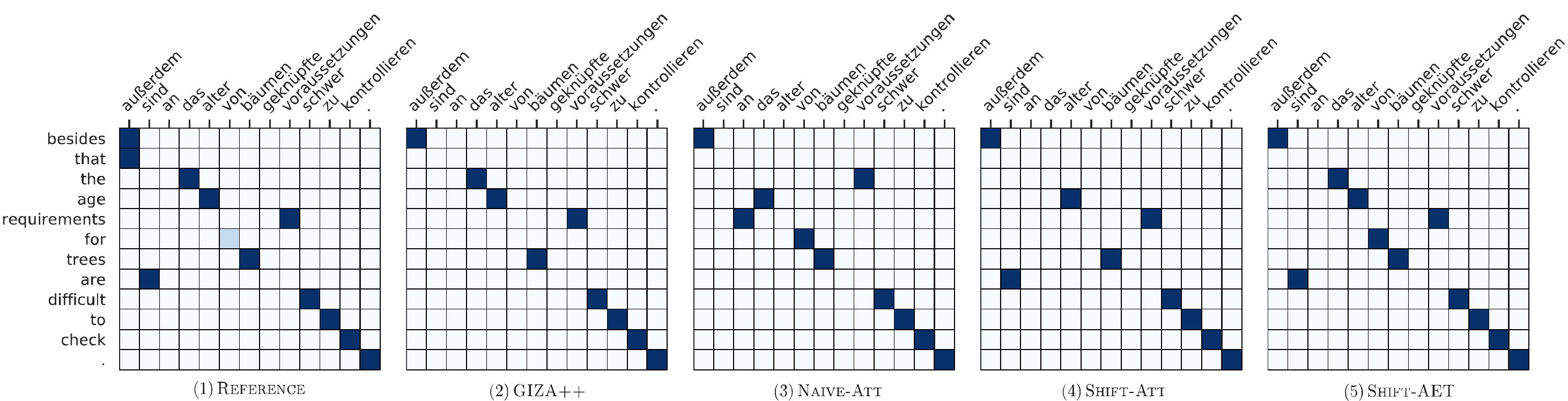}  %{figure/example3_short.jpg}
    \caption{One example from the de-en alignment test set. Golden alignments are shown in (1), blue squares and light blue squares represent \textit{sure} and \textit{possible} alignments separately. }
    % \vspace{-8pt}
    \label{fig:example1}
\end{figure*}
In Fig.~\ref{fig:example1}, we present a symmetrized alignment example from de-en test set. Manual inspection of this example as well as
others finds that our methods \textproc{Shift-Att} and \textproc{Shift-AET} tend to extract more alignment pairs than \textproc{GIZA++}, and extract better alignments especially for sentence beginning compared to \textproc{Naive-Att}. 

\section{Related Work}
Alignment induction from RNNSearch~\cite{Bahdanau2014NeuralMT} has been explored by a number of works. \citet{Bahdanau2014NeuralMT} are the first to show word alignment example using attention in RNNSearch. \citet{ghader-monz-2017-attention} further demonstrate that the RNN-based NMT system achieves comparable alignment performance to that of \textproc{GIZA++}. Alignment has also been used to improve NMT performance, especially in low resource settings, by supervising the attention mechanisms of RNNSearch \cite{chen2016guided,liu2016neural,alkhouli2017biasing}.

There is also a number of other studies that induce word alignment from Transformer. \citet{li-etal-2019-word,ding-etal-2019-saliency} claim that attention may not capture word alignment in Transformer, and propose to induce word alignment with prediction difference~\cite{li-etal-2019-word} or gradient-based measures~\cite{ding-etal-2019-saliency}. ~\citet{zenkel2019adding} modify the Transformer architecture for better alignment induction by adding an extra alignment module that is restricted to attend solely on the encoder information to predict the next word. \citet{garg-etal-2019-jointly} propose a multi-task learning framework to improve word alignment induction without decreasing translation quality, by supervising one attention head at the penultimate layer with \textproc{GIZA++} alignments. Although these methods are reported to improve over head average baseline, they ignore that better alignments can be induced by computing alignment scores at the decoding step when the to-be-aligned target token is the decoder input.
\section{Conclusion}
In this paper, we have presented two novel methods \textproc{Shift-Att} and \textproc{Shift-AET} for word alignment induction. Both methods induce alignments at the step when the to-be-aligned target token is the decoder input rather than the decoder output as in previous work. Experiments on three public alignment datasets and a downstream task prove the effectiveness of these two methods. \textproc{Shift-AET} further extends Transformer with an additional alignment module, which consistently outperforms prior neural aligners and \textproc{GIZA++}, without influencing the translation quality. To the best of our knowledge, it reaches the new state-of-the-art performance among all 
neural alignment induction methods. We leave it for future work to extend our study to more downstream tasks and systems.

\section*{Acknowledgments}
This work was supported by the National Key R\&D Program of China (No. 2018YFB1005103), National Natural Science Foundation of China (No. 61925601), the Fundamental Research Funds for the Central Universities and the funds of Beijing Advanced Innovation Center for Language Resources (No. TYZ19005). We thank the anonymous reviewers for their insightful feedback on this work.

% \clearpage
% \pagebreak

\bibliographystyle{acl_natbib}
\bibliography{emnlp2020}

\end{document}